# Real-time Multi-Object Tracking Based on Bi-directional Matching

Huilan Luo[1], Zehua Zeng[1*]

[1]Jiangxi University of Science and Technology, Ganzhou 341000, China

**Abstract**：In recent years, anchor-free object detection models combined with matching algorithms are used to achieve real-time muti-object tracking and also ensure high tracking accuracy. However, there are still great challenges in multi-object tracking. For example, when most part of a target is occluded or the target just disappears from images temporarily, it often leads to tracking interruptions for most of the existing tracking algorithms. Therefore, this study offers a bi-directional matching algorithm for multi-object tracking that makes advantage of bi-directional motion prediction information to improve occlusion handling. A stranded area is used in the matching algorithm to temporarily store the objects that fail to be tracked. When objects recover from occlusions, our method will first try to match them with objects in the stranded area to avoid erroneously generating new identities, thus forming a more continuous trajectory. Experiments show that our approach can improve the multi-object tracking performance in the presence of occlusions. In addition, this study provides an attentional up-sampling module that not only assures tracking accuracy but also accelerates training speed. In the MOT17 challenge, the proposed algorithm achieves 63.4% MOTA, 55.3% IDF1, and 20.1 FPS tracking speed.

**Keywords:** Multi-object tracking; Motion predictions; Realtime Tracking; Matching algorithms.

## 1. Introduction

The research of multi-target tracking can be traced back to the 1960s. Kalman filters are used to track radar-searched targets in earlier methods[1]. These methods are mostly tracking before detection and the targets are tracked based on radar signal characteristics[2]. Recently, tracking by detection is getting more and more popular due to the advanced performance of object detection based on deep networks. Tracking by detection first gathers the information of objects using the object detection algorithms and then completes the tracking using the matching algorithms. Multi-object tracking methods need to identify and track multiple objects in a video, and then assign consistent identity information to each specific object.

Depending on whether future frame information will be used as input to the algorithm, multi-object tracking methods can be divided into two modes: offline and online. The offline tracking mode[3]–[5] will first identify objects across the video and then link the identifications of each object and generate trajectories. On the other hand, the online tracking mode uses only the current frame and past frames for matching, so it can be used to analyze real-time video streams, allowing for wider applications. The proposed method in this article belongs to the online multi-object tracking.

Most of online multi-object tracking methods are tracking by detection[6], [7], which rely on object detection algorithms for object recognition and objection location. Then additional

---

convolutional neural network branches are used to obtain the features of detected objects. Therefore, the similarity between detected objects can be obtained based on their features. Besides, the Kalman filter algorithm[8] is usually used to estimate the motion information of objects. Finally, Hungarian matching algorithm[9] is used to match the objects between frames in terms of the obtained appearance similarity and motion information. Furthermore, recurrent neural networks have been employed in the research[10] to learn the temporal information of objects to improve the matching accuracy. More recently, some works[11]–[13] use sophisticated object detection architectures to obtain the outstanding tracking accuracy. In this paper, we focus on exploring effective and efficient matching algorithms without using appearance features of objects. Even with simple object detection models, our proposed algorithm can ensure fast training speed and online tracking.

The matching process usually necessitates the use of numerous pieces of information about the objects, such as object motion[14] and appearance features [15]. However, the extraction of appearance features is time-consuming, and it is difficult to achieve real-time object tracking. Some works try to remove appearance features and only use objects' motion estimation to improve tracking speed[16], but the matching can be easily interrupted when the object is occluded or disappears temporarily.

In order to solve the tracking interruption problem caused by object occlusions, a bi-directional tracking architecture is proposed in this paper, where two motion prediction branches are designed to obtain motion information about objects to get more accurate object location estimation for the subsequent matching process. The bi-directional motion predictions and object locations are used to match detected objects. After the first match is performed, the objects that are not matched will be temporarily stored in the stranded area. When they are detected again, the second match is performed. The proposed algorithm handles the object occlusion problem well without using appearance features of objects. Moreover, the attention mechanism is used in the up-sampling stage to optimize multi-level feature aggregations to learn more accurate locations of objects. The proposed attentional feature aggregation method attains significantly better training efficiency compared to the deformable convolution[17]. The main contributions of this paper are described briefly as follows:

(1) A bi-directional matching algorithm is proposed to effectively reduce tracking failures due to short-time occlusions of objects.

(2) An attentional up-sampling module is designed to optimize multi-level feature aggregations and improve training efficiency.

(3) A real-time bi-directional tracking architecture is constructed, which achieves 63.4% MOTA and 55.3% IDF1 in the MOT17 challenge, while running at 20.1 FPS tracking speed.

## 2. Related works

**2.1 Tracking-by-detection methods**

Detection-based multi-object tracking methods typically have three stages: (i) get the object position through the object detection network, (ii) extract the object's appearance features and motion information based on the object position, and (iii) feed the extracted information into a matching algorithm to obtain the final multi-object tracking results.

POI [18] is a representative tracking by detection model that uses the Faster R-CNN [19] as object detector, the GoogLeNet [20] to extract appearance features of object regions, and the

Kalman Filtering algorithm to predict object locations in the previous frame. After that, the Hungarian algorithm use both the object location information and the appearance information to do object matching to complement the tracking process.

Another impressing work SORT [16], which also uses Faster R-CNN as a detector, achieves real-time multi-object tracking by removing the appearance feature extraction process of object regions. But without using appearance features for matching results in some loss of the tracking performance. Therefore, the appearance feature extraction is re-introduced in DeepSORT [6], where ResNet [21] is used as the feature extraction network and a cascade matching algorithm is designed to balance tracking speed and accuracy.

However, even the sate-of-the-art object appearance feature-based tracking algorithm [22] can only track 0.7 frames per second, which is still insufficient for real-time tracking. The JDE [23] improves the tracking speed by integrating the feature extraction of objects into the object detection stage.

Most recently, several works [7], [24] use anchor-free object detection networks, such as CenterNet [25], as the backbone to obtain the centroid location and classification information of objects. The use of anchor-free object detectors for multi-object tracking can greatly reduce the complexity of the architecture and hence improve tracking speed. Furthermore, the CenterTrack [7] use CenterNet to predict the object positions, and based on this position information, it employs the greedy matching algorithm to obtain the associations between objects in two consecutive frames.

**2.2 Up-sampling optimization methods**

Anchor-based object detection algorithms [26] usually use the feature pyramid [27] to detect multi-scale objects by multi-level features. However, anchor-free object detection methods [28] are usually based on encoder-decoder architectures, where the encoded high-level features are decoded to high-resolution feature maps, based on which the locations, sizes and category information of objects are predicted.

Therefore, the decoding process, also known as up-sampling stage, is critical for anchor-free object detection models. During the stage, many methods fuse multi-level features from different layers of the encoder to optimize the up-sampling process to obtain good high-resolution semantic features, which will significantly improve the performance of the subsequent multi-object tracking.

To enhance the multi-scale feature fusion in the up-sampling stage, FairMOT [24] used the DLA-Net [29] as its backbone network. During the decoding, where a tree-like structure is designed to iteratively up-sample the feature maps by fusing features from shallower layers of the encoder. CornerNet [30] uses the Hourglass network [31] as its backbone network, which is commonly used for human key point detection. Its multiple down-sampling and up-sampling process allows it to better fuse multi-scale features. However, the repeated stacking of hourglass structure makes the network too deep to be applied to real-time multi-object tracking.

Deformable convolution (DCN) [17] is used in the up-sampling process of recent anchor-free object detection algorithms and multi-object tracking algorithms [31], [32] to fuse multi-scale information and increase tracking accuracy. However, each convolutional kernel of the deformable convolution has its own offset, making it difficult to optimize in parallel and ultimately increasing the model training time significantly. Other researchers have applied up-sampling methods used in semantic segmentation, such as dilated convolution [33], to multi-object tracking [34], which can bring a little improvement of the detection accuracy.

In this paper, we propose an attentional up-sampling structure to boost attention to object location and motion information [35], [36], therefore improve the final detection accuracy and tracking performance.

**2.3 Matching algorithms**

The matching algorithm is also very important for the performance of multi-object tracking by detection models. Given object detection results, a matching algorithm links the identical objects between the consecutive frames. It can be abstracted as a maximum weighted matching problem, where the weights are the similarities between the targets in two frames. After generating the weighted bipartite graph, the object tracking is completed by finding the maximization assignment.

The traditional multi-object tracking algorithms [6], [16] use Kalman filtering to anticipate the positions of objects in the current frame based on their locations in the previous frame. The location distance and/or the appearance similarity between objects are used as the weights of the edges of the weighted bipartite graph. Based on the graph, the Hungarian method [9] is often used to complete the object matching.

On the other hand, CenterTrack [7] uses backward prediction to infer the object positions in the previous frame based on information from the current frame, and then greedy matching is carried out according to the target location and object confidence, in which the nearest two targets are considered to be the same target.

During object tracking, objects may disappear due to occlusion or detection failures, but these lost objects may reappear in future frames. Because the online algorithm does not use future frames and only uses targets detected in the current and previous frames for matching, object occlusion usually leads to tracking interruptions.

In order to address this problem, DeepSORT [6] proposes the cascade matching to provide a second match, which helps to reduce track interruptions caused by object occlusions. All detected bounding boxes are characterized as trajectory boxes, which assigns a matching priority based on the number of successful matches and failed matching rounds. Firstly, the trajectory box with high priority uses its position information for Hungarian matching. Then, the low priority trajectory box will use its appearance information to perform a second match with the previously unmatched boxes. The cascade matching strategy can minimize the interruption of trajectories. Similar matching algorithms have been used in many subsequent studies [37], [38].

This paper proposes a greedy cascade matching algorithm based on bi-directional motion prediction information, which effectively solves the tracking interruption problem caused by object occlusion and improves the tracking performance by using backward motion prediction information in the first matching and forward motion prediction information in the second matching.

**3. Model design for bi-directional tracking**

The bi-directional tracking model proposed in this paper is illustrated in Figure 1. It is composed of six parts, which are the input stage, the down-sampling stage, the up-sampling stage, the output stage, the first matching stage and the second matching stage.

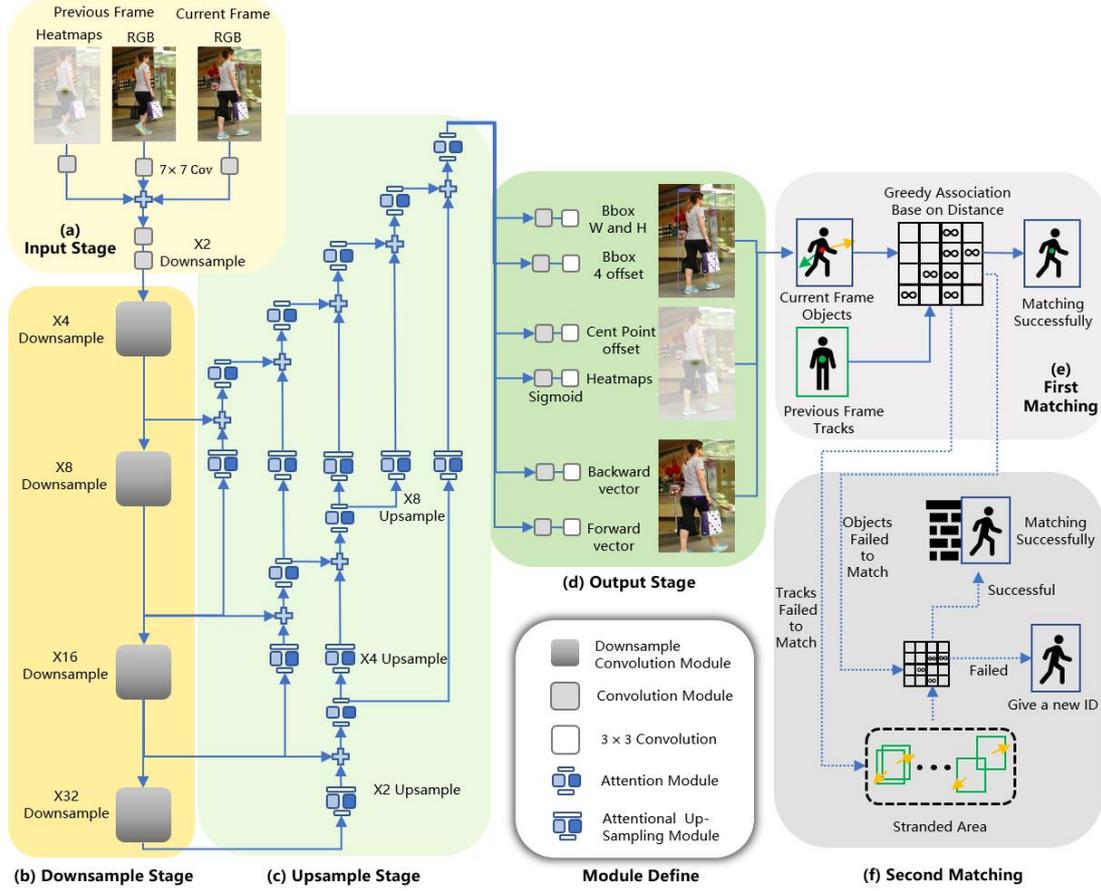

Fig.1 Bi-directional real time tracking network

## 3.1 Architecture Overview

The network inputs include the RGB images of the current and previous frames, and the heatmap of the previous frame generated by the model. These three inputs are expressed in Eq. (1) as $x_1, x_2, x_3$. They are individually processed through a $7\times 7$ convolution layer for initial feature extraction, as illustrated in Figure 1(a), and then summed to generate the feature maps with 16 channels. The feature maps are then down-sampled twice using two $3\times 3$ convolutional layers to generate the feature maps with 32 channels. Eq. (1) depicts this process, with $Cov$ denoting the convolution operation and its subscript $i$ denoting the convolution stride size.

$$F(x_1, x_2, x_3) = CovBR_2\left(CovBR_1\left(\sum_{i=1}^{3} CoveBR_1(x_i)\right)\right)$$
$$CovBR_i(x) = \mathrm{ReLU}(Batchnorm(Cov_i(x)))$$
(1)

Then the obtained initial features are transmitted to DLA-Net [29] for further feature extraction. As shown in Figure 1(b), the DLA-Net down-sampling network has four down-sampling convolutional blocks, each of which completes a $2\times$ down-sampling and doubles the number of channels. Next, the multi-scale feature maps generated by these four blocks are sent into the proposed attentional up-sampling module for feature fusion, and we get the final feature maps with the size of 1/4 of the original input image size, as depicted in Figure 1(c).

As shown in Figure 1(d), the feature maps obtained through attentional up-sampling will be fed into six branches to learn heatmaps, center point offsets, object widths and heights, offsets of the four edges of the bounding box from the center point, and bi-directional motion vectors. These outputs are used in the following two matching stages.

The first matching stage utilizes the motion vectors and other object detection information to predict the positions of the objects in the previous frame. Based on these object locations, we can get the distance matrix between objects of the two different frames. With the distance matrix and the object confidence scores, the greedy matching algorithm is employed to complete the first matching, as illustrated in Figure 1(e). In this matching process, two types of objects may fail to match. The first is that the object in the current frame is occluded and the other is that the object is newly-occurred or it was occluded in the previous frame.

To track the temporally occluded objects successfully, we proposed the second matching stage, where a stranded area is designed to store the failed tracks, as illustrated in Figure 1(f). When targets reappear in a future frame, it can be matched with the tracks in the stranded area by the second matching algorithm. Only the object that fails to match in the second matching stage, it will be considered as a newly emerged object and be assigned a new ID.

**3.2 Attentional up-sampling**

The proposed attentional up-sampling stage based on DLA network [29] focuses on adaptively fusing multi-level features to effectively and efficiently decode features to get more accurate object information. As shown in Figure 1(c), the higher-level features are up-sampled using the attentional up-sampling module before it is aggregated with the consecutive lower-level features. After using an attentional module to compress the numbers of channels, the integrated features are again up-sampled with the attention mechanism and then fused with the next-level aggregated features. The iterative attentional aggregation method allows the feature fusion process to progress from shallow to deep, resulting in a high-resolution feature map.

The attention-based up-sampling strategy utilized in this paper boosts region-specific information and speeds up the training of the model. By adding the spatial attention into the up-sampling process, the saliency of the object position information can be enhanced, resulting in improved detection accuracy of object center position. Considering the number of channels will be greatly reduced during the up-sampling process, we also leverage the channel attention to effectively improve feature screening to improve the model's multi-output performance.

The attention mechanism [36] used in the up-sampling stage is illustrated in Figure 2. The channel attention enhancement and spatial attention enhancement are carried out successively. Then a convolutional layer is used to implement the adjustment of the numbers of channels. For up-sampling, the deconvolution layer is followed to double the resolution of feature maps.

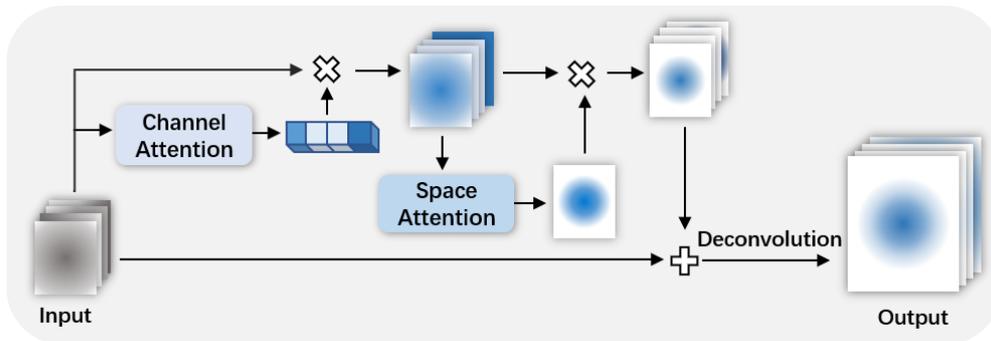

Fig. 2  Attentional up-sampling module

The channel attention uses the channel compression excitation method. The process is shown in Eq. (2), where $Cov$ denotes a $1\times 1$ convolution operation that enables information communication between dimensions by reducing dimensions first and then increasing it.

$$F(x) = Sigmoid(CovR(MaxPooling(x)) + CovR(AvgPooling(x)))$$
$$CovR(x) = cov(ReLU(cov(x))) \quad (2)$$

Spatial attention is implemented by first using global average pooling and global maximum pooling respectively along the channel dimension, and then concatenating their results. Next, a $7\times 7$ convolution layer to fuse the two channels into one channel and a Sigmoid function to generate the final attention weights on spatial dimensions. The process is shown in Eq. (3).

$$F(x) = Sigmoid(cov(Concatenate(MaxPooling(x), AvgPooling(x)))) \quad (3)$$

**3.3 Six output branches**

The six output branches can be divided into three pairs: (i) to learn the heatmap of object locations and its offset, (ii) to learn object size and its offset, and (iii) to learn the bi-directional motion information of objects, e.g., relative to the future frame and the previous frame respectively. Each branch first uses a $3\times 3$ convolution layer to increase the number of channels from 64 to 256, and then uses a $3\times 3$ convolution layer to further learn the specific task and modify the final number of channels to suit the various outputs. A Sigmoid function has been added to the heatmap branch to provide probabilistic information on the location of center points. A target is considered to exist when a pixel above a threshold value appears in the heatmap. The corresponding object scales and object movement vectors can be obtained from other branches based on the object heatmap.

**The heatmap and its offset**. When training the heatmap branch, its label is obtained by formalizing the center point locations of objects as a two-dimensional normal distribution, as shown in Eq. (4).

$$G(x,y) = \frac{1}{2\pi r^2} e^{\frac{-(x^2+y^2)}{2r^2}} \quad (4)$$

Where $(x,y)$ denotes the center point and the radius $r$ is obtained by solving the equation shown in Eq. (5).

$$w \times h \times rate = \left(w - \frac{h*r}{\sqrt{w^2+h^2}}\right) * \left(h - \frac{w*r}{\sqrt{w^2+h^2}}\right) \quad (5)$$

Where $w$ and $h$ denote the object bounding box's width and height, and $rate$ is a hyperparameter, which is set to 0.7 in our experiments.

For the heatmap, because most of the points are negative samples, we use Focal Loss [39] as the loss function to alleviate the uneven sample problem, expressed as Eq. (6).

$$L_{heat} = \frac{1}{N} \sum_{x,y} \begin{cases} (1-\hat{Y})^\alpha \log(\hat{Y}) & Y=1 \\ (1-Y)^\beta (\hat{Y})^\alpha \log(1-\hat{Y}) & Y=0 \end{cases} \quad (6)$$

Where $\hat{Y}$ represents the predicted probability of the object appearing at $(x,y)$ and $Y$

represents the ground truth. $N$ represents the total number of detected objects.

The maximum pooling is used to extract the peak of the heatmap, which is then filtered by a confidence threshold to produce the initial center locations of objects. Then, the output of the center point offset branch is added to it to get an accurate object center location.

**Object size and its offset branches.** Two branches are designed to anticipate the object scale information. The first branch estimates the object height and width, while the second branch predicts the distance from the object's center to the four edges of its bounding box. Because sometimes part of an object will move out of the frame, the bounding box does not completely cover the entire object. Therefore, by combining these offset information, more accurate size of objects can be obtained to improve the performance of multi-object tracking.

**Bi-directional motion prediction branches.** In order to address the tracking interruption problem because of object occlusions, two branches are designed to learn the bi-directional motion information. One is to estimate the motion vector relative to the previous frame, which is utilized in the first matching stage to calculate its location in the previous frame. In the second matching stage, another motion vector relative to the future frame is used by the occluded targets in the strand area to predict their locations in the future frames, so they can be matched successfully when they appear again.

The ground truth of motion vectors can be computed directly from the object trajectory when trained with the multi-object tracking dataset. When the object detection dataset is used for pre-training, the objects are translated to simulate their movements.

**Loss function.** The total loss function is a weighted sum of the losses from the six branches, as shown in Eq. (7), where $L_{heat}$ denotes the loss for the heatmap branch, $L_{cent\ offcet}$ for the object center offset branch, $L_{wh\ offcet}$ for the width and height of the object bounding box, $L_{edge\ offcet}$ for the offset of the four edges of the bounding box relative to the center, $L_{backward}$ and $L_{forward}$ for the motion vectors pointing to the previous and future frames respectively. Except the heatmap branch, where Focal Loss [39] is used, all the other branches are trained using the Smooth L1 [40] loss function. Because the scale of object sizes is very much larger compared with that of the motion vectors and other outputs, their losses are weighted by 0.1.

$$L_{sum} = L_{heat} + L_{cent\ offcet} + 0.1 L_{wh\ offcet} + 0.1 L_{edge\ offcet} + L_{backward} + L_{forward} \quad (7)$$

## 4. Bi-directional matching algorithm

This research presents a bi-directional matching algorithm to improve the tracking performance under the scenarios of occurring object occlusions. The proposed algorithm performs two distance-based greedy matching processes by utilizing the bi-directional motion information and a stranded area. The tracking boxes that are failed to match in the first matching stage are moved to the stranded area, where their positions keep updating in in terms of the object motion prediction. Therefore, when the objects appear again, they can be matched in the second matching stage.

### 4.1 Distance-based greedy matching

Because the motion of objects in video is basically continuous, a pair of objects close to each other between frames are more likely to be the same target. Based on this assumption, the matching method based on distance is adopted, which greatly improves the tracking efficiency and realizes real-time tracking.

The distance-based matching is relied on the calculation of the distance matrix between objects in the previous frame and the current frame. The motion vector relative to the previous frame, e.g., backward motion vector, is used on the detected objects to predict their locations in the previous frame. Therefore, based on the predicted positions, the distance matrix between objects in the two frames can be obtained as the Algorithm 1.

When the distance between two objects is larger than one of their bounding box area, it is changed to the infinity, which means that the two objects cannot match each other. After obtaining the distance matrix, the detected targets with high confidence scores are matched first to the closest objects.

---

**Algorithm 1 Distance matrix calculation**

**Input**: object center point $C_t$ of the previous frame; the area $A_t$ of the object bonding box of the previous frame; object center point $C_d$ of the current frame; the area $A_d$ of object bonding box of the current frame; and object movement prediction $M_d$

**Output**: $Distance$

1. $C'_d = C_d + M_d$
2. $Distance = (C'_d - C_t)^2$
3. **For** each element $(t, d)$ **in** $Distance$ **Do**
4.     **If** $Distance(t, d) > Min(A_t, A_d)$ **Then**
5.         $Distance(t, d) = \infty$
6.     **End If**
7. **End For**

---

**4.2 The proposed bi-directional Matching algorithm**

The proposed bi-directional matching algorithm consists of two matching stages, both of which are based on the distance-based greedy matching strategy. In the first matching stage, the successfully matched objects will inherit the ID information from the previous frame. But the objects that are not matched will be saved to the strand area to wait the chance of the second matching.

There are two situations that can lead to the first matching failure: one is that there are newly appeared objects in the current frame, the second is that the objects once occluded in the previous frame appear again. To avoid mistaking the re-appear objects as new objects, which is the main challenge for the multi-object tracking, we created a stranded area to temporarily hold occluded objects to allow them to have chance to do the second matching when they appear again. In order to use the distance-based greedy algorithm, the target in the stranded area should be continually moved by using the motion vector relative to the future frame, simulating the movement of a target behind an occlusion in the actual world. Therefore, when the object reappears, they can be linked by the distance-based greedy matching algorithm, to maintain the tracking trajectory's continuity. Figure 3 depicts the procedure.

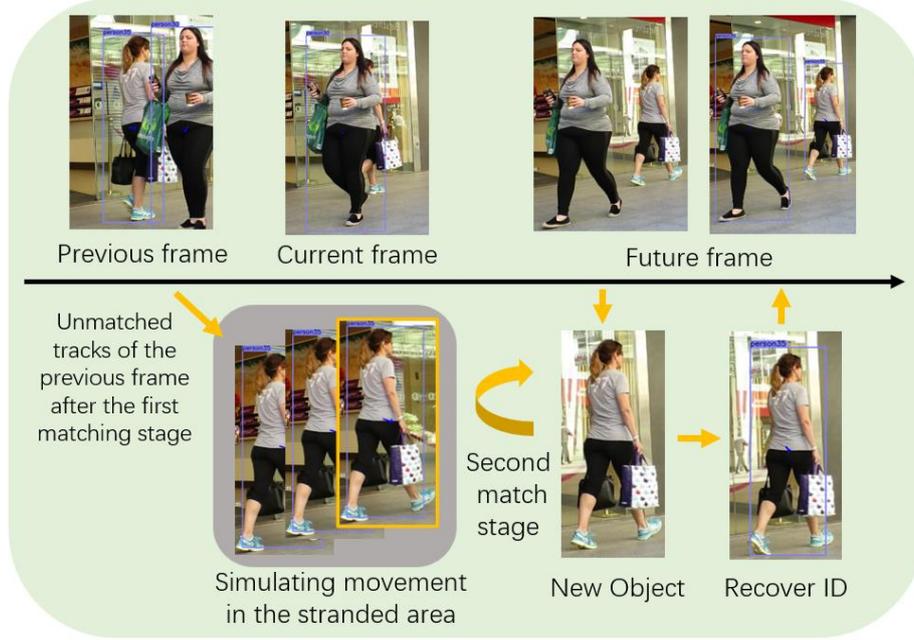

Fig.3 Stranded area operation process

How long the object should stay in the stranded area depends on when the subject reappears. There is a situation where the object may no longer return. Therefore, we need to delete these objects from the stranded area. In order to identify which objects may no longer return, the objects are given a life value when they enter the stranded area, and the score is reduced over time. The objects will be removed when their life values become 0.

The overall matching algorithm is illustrated below in Algorithm 2. If the unmatched objects in the first matching stage cannot match successfully with the objects in the stranded area in the second matching stage, it is assigned a new ID and considered as a new object.

---

**Algorithm 2 Bi-directional matching algorithm**

**Input:** the track indices in the previous frame $T = \{1,...,N\}$, The detection indices $O = \{1,...,M\}$ sorted by their confidence scores, Stranded area $S = \{...\}$ initialized with the detections of the very first frame, the life values $L = \{l_i\}$, Motion vectors towards future frames $V = \{v_i\}$, Center points of objects $C = \{c_i\}$.

**Output:** matching result A

1. Initialize set of matches $A = \varnothing$
2. Compute the distance matrix $D_1$ between the tracks and the detections using Algorithm 1
3. **For** $i = 1$ to $\min(N, M)$
4.     $j = arg\min_k D_1(i,k)$
5.     **If** $D_1(i,j) \neq \infty$
6.         $A = A \cup \{(i,j)\}$
7.         $T = T - \{j\}$
8.         $O = O - \{i\}$
9.     **End If**
10. **End For**
11. **If** $O \neq \varnothing$
12.     Compute the distance matrix $D_2$ between the objects in the stranded area and the detections in $O$ using Algorithm 1

|   |   |   |
|---|---|---|
| 13. | | **For** $i$ in $O$ |
| 14. | | $j = arg\min_{k} D_2(i,k)$ |
| 15. | | **If** $D_2(i,j) \neq \infty$: |
| 16. | | $A = A \cup \{(i,j)\}$ |
| 17. | | $O = O - \{i\}$ |
| 18. | | $S = S - \{j\}$ |
| 19. | | **End If** |
| 20. | | **End For** |
| 21. | **End If** | |
| 22. | Update the life values of objects in the stranded area: $l_i = l_i - 1$ | |
| 23. | Remove the objects from $S$ if the life value is zero. | |
| 24. | Add the unmatched tracks into the stranded area and initialize their life value: $S = S \cup T$, $l_i = L_{\max}$ $for$ $i \in T$ | |
| 25. | Update the locations of the objects in $S$: $c_i = c_i + v_i$ | |
| 26. | **Return** $A$ | |

## 5. Experimental results and analysis

### 5.1 Performance Measures

There are six mainstream metrics for multi-object tracking: IDs, FM, MOTA [41], IDF1 [42], MT and ML.

IDs (Identity Switches) represents the ratio of the number of ID changes to the number of ground truth objects.

FM (Fragmentation) denotes the ratio of the number of times the tracking trajectory is interrupted to the number of ground truth objects.

MOTA (Multiple Object Tracking Accuracy) is calculated as shown in Eq. (8), where $t$ is the frame index, FN is the number of missed targets, FP is the number of false positive objects and GT is the number of ground truth objects. It is an overall metric. However, detection performance has a greater impact on MOTA than tracking performance.

$$MOTA = 1 - \frac{\sum_t (FN_t + FP_t + IDs_t)}{\sum_t GT_t} \tag{8}$$

IDF1 is the identification F1 score. It is calculated as shown in Eq. (9), where IDTP indicates the number of correctly matched Identifications, while IDFP and IDFN indicate the number of incorrectly matched and unmatched Identifications, respectively.

$$IDF_1 = \frac{2 \times IDTP}{2 \times IDTP + IDFP + IDFN} \tag{9}$$

MT (Mostly Tracked) indicates the ratio of mostly tracked targets to the total number of ground truth trajectories. A target is mostly tracked if it is successfully tracked for at least 80% of its life span.

ML (Mostly Lost) denotes the ratio of mostly lost targets to the total number of ground truth trajectories. If a track is only recovered for less than 20% of its total length, it is said to be mostly lost.

## 5.2 Dataset and experiment setups

Currently, the mainstream benchmarks for the multi-object tracking research are the MOT Challenge series [43], including MOT15, MOT16, MOT17, etc. The MOT17 dataset is used in this paper to train and test our method. MOT datasets focus on the pedestrian tracking and cover a variety of tracking scenarios, including fixed camera tracking, mobile camera tracking, night tracking, and indoor tracking. There are lots of object occlusions and great object appearance changes in MOT17, which pose great challenges to MOT algorithms.

In the following experiments, for our method, DLA-34 pre-trained on ImageNet is used as the backbone network. The crowd-human dataset [44] is also used to train the detection model, where the dataset is augmented with random cuts to simulate object displacements. The batch size is set to 18. The $\alpha$ and $\beta$ are set to 2 and 4 for the Focal Loss [39]. Because more than 70% of the occlusion durations in the MOT17 dataset are less than 20 frames, the initial life value of the objects in the stranded area is set to 20.

## 5.3 Ablation experiment for the attentional up-sampling stage

This experiment uses the basic tracking algorithm CenterTrack [7] to evaluate the effectivity and efficiency of the proposed attentional up-sampling method. We compared four different up-sampling methods: (i) ordinary convolution, (ii) deformable convolution [17], (iii) an asymmetric convolution model [45], and (iv) the attentional up-sampling method proposed in this paper.

The tracking results of the four methods on the MOT17 dataset are shown in Table 1. From the experimental results, the attentional up-sampling method proposed in this paper achieves much better performance than other methods in terms of the IDF1, MT and ML. And it very closes to the first place in terms of MOTA, IDs and FM. Therefore, the experimental results show that the attentional up-sampling method has a good impact on the tracking performance.

Table 1 Performance comparison of multiple up-sampling methods

| Model | MOTA %↑ | IDF1 %↑ | IDs %↓ | FM %↓ | MT %↑ | ML %↓ |
|---|---|---|---|---|---|---|
| Normal Convolution | 48.0 | 50.1 | 1.4 | 2.0 | 25.4 | 31.3 |
| Deformable Convolution | **49.9** | 53.6 | 1.2 | 1.8 | 24.8 | 32.2 |
| Asymmetric Convolution | 48.8 | 52.6 | **1.0** | **1.5** | 25.1 | 33.3 |
| **Attention Model** | 49.8 | **56.2** | 1.1 | 1.7 | **27.4** | **28.9** |

The number of parameters and training speed of the four up-sampling methods are also compared in Table 2. It can be seen that the attentional up-sampling method has fewer parameters and faster training speed compared to the deformable convolutional up-sampling and the asymmetric convolution method. The proposed attentional up-sampling method significantly improves the tracking performance with minimal increase of computational cost compared with the normal convolution method.

Table 2 comparison of parameter quantity and training speed of multiple up-sampling methods

| Model | model parameters | 1 epoch of training time (minutes) |
|---|---|---|
| Normal Convolution | **16.91M** | 1 |
| Deformable Convolution | 20.47M | 3.86 |
| Asymmetric Convolution | 22.10M | 1.21 |
| **Attention Model** | **19.86M** | 1.13 |

**5.4 Comparisons of the Bi-directional tracking algorithm with the baseline**

By designing a stranded area to temporally store the unmatched tracks and simulating the movement according to the forward motion vector, the proposed bi-directional tracking algorithm can reduce object identity switching caused by detection fluctuations or object occlusions, thereby improving tracking quality. In order to verify the effectiveness of the proposed method, we conducted three experiments: (i) comparing tracking performance with the baseline method CenterTrack, (ii) comparing the training process, and (iii) comparing the effectiveness of addressing the occlusion problem.

The tracking results of the bi-directional tracking method and the baseline CenterTrack are reported in Table 3. For fairness, these models are trained and tested under the same conditions. From the experimental results in Table 3, it can be seen that the bi-directional tracking model has indeed reduced identity switching according to IDF1 and IDs.

Table 3 Performance comparison with improved models and algorithms

| Model | MOTA %↑ | IDF1 %↑ | IDs %↓ | FM %↓ | MT %↑ | ML %↓ |
|---|---|---|---|---|---|---|
| CenterTrack | 59.7 | 61.0 | 1.2 | 1.7 | 36.9 | 21.1 |
| **Our method** | **60.3** | **61.5** | **0.9** | **1.6** | 36.6 | 22.4 |

In order to clearly illustrate that how effective our tracking method is in reducing the IDs. Figure 4 shows a specific object tracking details when the object is occluded and then re-appear, where target-65 is almost completely obscured by target-43 in the middle frame. As shown in the first row of Fig. 4, CenterTrack give it a new ID 109 when target-65 re-appear. However, as shown in Figure 4(b), our bi-directional tracking algorithm can temporarily store the occluded target-65 in the stranded area and recover its ID when it reappears, realizing uninterrupted tracking.

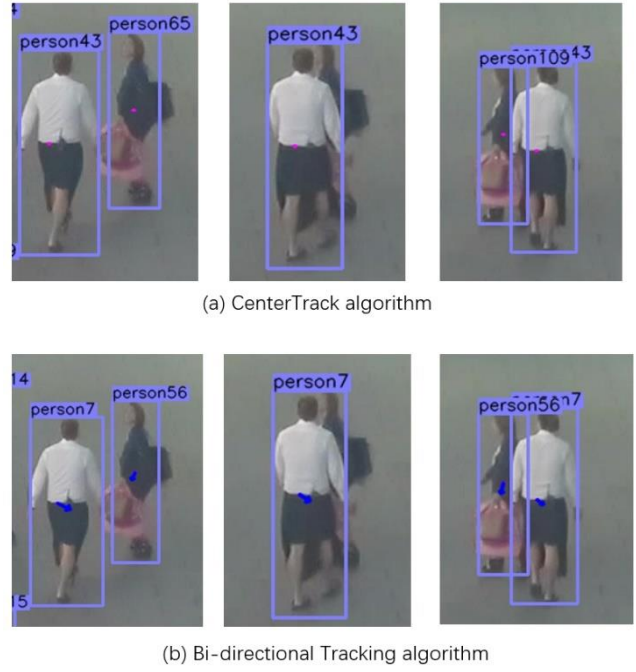

Fig.4 An example of tracking an occluded object: (a) CenterTrack; (b) our bi-directional tracking algorithm; where the blue arrow denotes the backward motion vector and the red arrow denotes the forward motion vector.

Additionally, Fig.5 shows the Training and IDF1 values during training of using the proposed bi-directional matching algorithm compared to the baseline matching algorithm: distance-based greedy matching algorithm. As seen in Fig. 5, our method keeps better tracking performance during the whole training process. It demonstrates that the bi-directional matching algorithm is able to provide IDF1 gains no matter how well the detector trained.

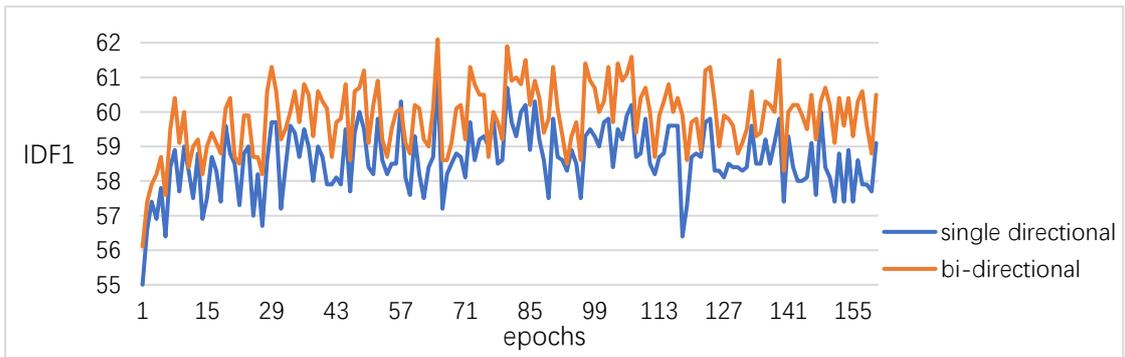

Fig.5 Training and IDF1 values during training between epochs 1 and 160, of the bi-directional matching method compared to the baseline

In order to further verify the ability of the bi-directional tracking method to solve the object occlusion problem, we first change the MOT17 dataset by randomly masking different pedestrians to simulate occlusion situations with different occlusion rates, as shown in Figure 6. Then we compare the performance on the changed MOT dataset of two methods: the traditional single-direction tracking method used in CenterTrack and the proposed bi-directional matching algorithm. For fairness, both of them are based on the same detection architecture described in section 3. The results are shown in the Table 4. It can be seen from the experimental results that the performance gain of the bi-directional tracking algorithm is increasing with the increase of occlusion rate. This demonstrates that the bi-directional tracking algorithm is effective to solve the occlusion or missing detection problem, and therefore it can improve the tracking robustness and tracking accuracy.

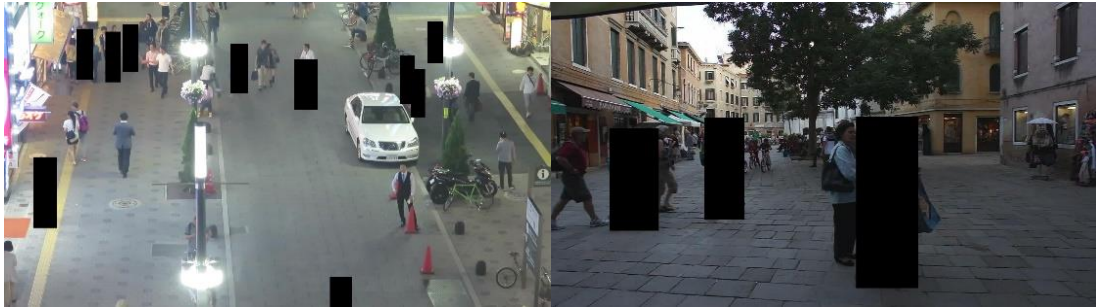

Fig. 6 Illustration images of random occlusions

Table 4 Comparison of algorithm performance on the occluded MOT17 dataset

| Occlusion rate | MOTA(%↑) | | IDF1(%↑) | | IDs(%↓) | |
| --- | --- | --- | --- | --- | --- | --- |
| | Single directional | **Bi-directional** | Single directional | **Bi-directional** | Single directional | **Bi-directional** |
| 0% | 60.25 | **60.27** | 60.2 | **61.5** | 1.04 | **0.88** |
| 5% | 56.68 | **57.32** | 35.9 | **47.9** | 2.84 | **2.02** |
| 10% | 50.39 | **51.86** | 23.9 | **35.7** | 6.10 | **4.37** |
| 15% | 43.00 | **45.85** | 14.7 | **26.2** | 9.36 | **6.20** |
| 20% | 36.98 | **40.20** | 10.9 | **20.5** | 11.42 | **7.83** |
| 25% | 31.87 | **35.84** | 9.1 | **18.3** | 12.91 | **8.62** |
| 30% | 28.12 | **32.33** | 8.3 | **15.8** | 13.57 | **9.07** |

**5.5 Comparison with advanced methods**

Table 5 presents the results of our method and other high-performance models of recent years. The results come from the officially published models on the MOT Challenge leader board. Although our method is online, we also compare some results of the best performing offline methods to highlight competitive performance of our approach. Our work and GSDT [46] significantly outperforms other MOT methods in most of metrics. GSDT uses an additional graph neural network to improve matching accuracy, which makes the tracking speed slower than our work. It also should be noted that 6 pedestrian detection datasets are utilized in GSDT for pre-training to get the best performance, but our work only uses one.

Compared with other state-of-the-art MOT methods with large detection models (TubeTK) [45] or sophisticated object association processing (TPM [47], SST [22]). Although our work uses the simpler detection model, it outperforms TPM in all evaluation metrics and surpasses TubeTK and SST in most metrics.

Overall, by comparing our work with other multi-object tracking algorithms, our work is faster and better, and outperforming most algorithms of the past two years in terms of overall performance, demonstrating the effectiveness and efficiency of our approach.

Table 5 Comparisons of our method with state-of-the-art models on MOT17 dataset.

| Algorithm | Year | Type | MOTA %↑ | IDF1 %↑ | MT %↑ | ML %↓ | FP ↓ | FN ↓ | Speed (FPS)↑ |
|---|---|---|---|---|---|---|---|---|---|
| Tracktor++[14] | 2019 | Online | 56.3 | 55.1 | 21.1 | 35.3 | **8866** | 235449 | 1.5 |
| TPM[47] | 2020 | Offline | 54.2 | 52.6 | 22.8 | 37.5 | 13739 | 242730 | 0.8 |
| BLSTM[48] | 2021 | Online | 51.5 | 54.9 | 20.4 | 35.5 | 29616 | 241619 | **20.1** |
| LPC_MOT[49] | 2021 | Offline | 59.0 | 66.8 | 29.9 | 33.9 | 23102 | 206948 | 4.8 |
| SST [22] | 2019 | Online | 52.4 | 49.5 | 21.4 | 30.7 | 25423 | 234592 | 6.3 |
| TubeTK [45] | 2020 | Online | **63.0** | 58.6 | **31.2** | 19.9 | 27060 | **177483** | 3 |
| GSDT[46] | 2021 | Online | **66.2** | 68.7 | **40.8** | 18.3 | 43368 | **144261** | 4.9 |
| **Ours** | 2021 | Online | **63.4** | 55.3 | **33.2** | 27.3 | 26538 | 173172 | **20.1** |

# 6. Conclusion

In this paper, we propose a bi-directional tracking algorithm that makes full use of bi-directional motion estimation, combined with distance-based greedy matching and a stranded area, to alleviate tracking interruptions due to short-time occlusion. Our proposed approach improves the tracking performance in the case of object occlusion and achieves good results at real-time tracking speed on the MOT17 test set. Moreover, the proposed attentional up-sampling module is also experimentally approved to be effective and efficient. Further work will focus on improving the accuracy of object motion estimation and solving the problem of uneven distribution of motion vector training samples.